\documentclass[10pt,twocolumn,letterpaper]{article}

\usepackage{cvpr}
\usepackage{times}
\usepackage{url}            
\usepackage{booktabs}       
\usepackage{amsfonts}       
\usepackage{nicefrac}       
\usepackage{microtype}      
\usepackage[mathscr]{euscript}
\usepackage{graphicx}
\usepackage{amsmath}

\usepackage{arydshln}
\usepackage{float}
\restylefloat{table}

\usepackage[pagebackref=true,breaklinks=true,colorlinks,bookmarks=false]{hyperref}

\makeatletter
\def\adl@drawiv#1#2#3{%
        \hskip.5\tabcolsep
        \xleaders#3{#2.5\@tempdimb #1{1}#2.5\@tempdimb}%
                #2\z@ plus1fil minus1fil\relax
        \hskip.5\tabcolsep}
\newcommand{\cdashlinelr}[1]{%
  \noalign{\vskip\aboverulesep
           \global\let\@dashdrawstore\adl@draw
           \global\let\adl@draw\adl@drawiv}
  \cdashline{#1}
  \noalign{\global\let\adl@draw\@dashdrawstore
           \vskip\belowrulesep}}
\makeatother

\cvprfinalcopy 

\def\cvprPaperID{7005} 

\ifcvprfinal\fi

\title{HUSE: Hierarchical Universal Semantic Embeddings}

\author{Pradyumna Narayana$^1$, Aniket Pednekar$^1$, Abishek Krishnamoorthy\thanks{This work was conducted at Google.}    $^{\, 2}$, Kazoo Sone$^1$, Sugato Basu$^1$\\$^1$Google, $^2$Georgia Institute of Technology\\{\tt\small \{pradyn,aniketvp,sone,sugato\}@google.com, akrishna61@gatech.edu}}

\begin{document}

\maketitle
\begin{abstract}
There is a recent surge of interest in cross-modal representation learning corresponding to images and text. The main challenge lies in mapping images and text to a shared latent space where the embeddings corresponding to a similar semantic concept lie closer to each other than the embeddings corresponding to different semantic concepts, irrespective of the modality. Ranking losses are commonly used to create such shared latent space --- however, they do not impose any constraints on inter-class relationships resulting in neighboring clusters to be completely unrelated. The works in the domain of visual semantic embeddings address this problem by first constructing a semantic embedding space based on some external knowledge and projecting image embeddings onto this fixed semantic embedding space. These works are confined only to image domain and constraining the embeddings to a fixed space adds additional burden on learning.
This paper proposes a novel method, HUSE, to learn cross-modal representation with semantic information. HUSE learns a shared latent space where the distance between any two universal embeddings is similar to the distance between their corresponding class embeddings in the semantic embedding space. HUSE also uses a classification objective with a shared classification layer to make sure that the image and text embeddings are in the same shared latent space.  Experiments on UPMC Food-101 show our method outperforms previous state-of-the-art on retrieval, hierarchical precision and classification results.
\end{abstract}

\section{Introduction}

The internet has been evolving from primarily being text-based to being multimodal, where text content is augmented with content from image or video modalities. With the rapid growth of different media types, learning discriminative feature representations of all these modalities is an essential yet challenging problem. One key useful feature is that when data from different modalities share the same semantics or have latent correlations (e.g., picture of baby and audio of a baby crying), we would like the resulting multimodal embeddings to share a common latent space. This space would allow the embeddings from multiple modalities to exploit the complementary information among themselves, while enriching the resulting common latent space. The resulting modality-agnostic (universal) embeddings would highly benefit 
search, ranking, ads and e-commerce space. Moreover, such universal embeddings are highly relevant for cross-modal retrieval. However, learning universal embeddings is a challenging task due to the ``media gap"~\cite{peng2017overview}, which means features from different modalities can be inconsistent.

There is significant recent work in the domain of cross-modal retrieval for creating a universal embedding space. Majority of these approaches create a latent space by aligning the embeddings from different modalities using ranking losses~\cite{ye2018advise, wang2016learning, kiros2014unifying, faghri2017vse++, carvalho2018cross, zhen2019deep, sarafianos2019adversarial}. These losses allow the embeddings corresponding to a semantic class to lie closer to each other than the embeddings corresponding to two different classes. However, they do not impose any constraints on inter-class relationships. So, the neighboring clusters may be completely unrelated resulting in the learned universal embedding space to not be semantically meaningful~\cite{barz2019hierarchy}. For example, the average embedding distance between \textit{cat} and \textit{dog} classes might be as large as the average distance between \textit{cat} and \textit{bridge} classes. However, this behavior is not desirable in retrieval problems where it is more meaningful to get semantically similar results.

\begin{figure}[ht]
\includegraphics[width=\linewidth, height=0.8\linewidth]{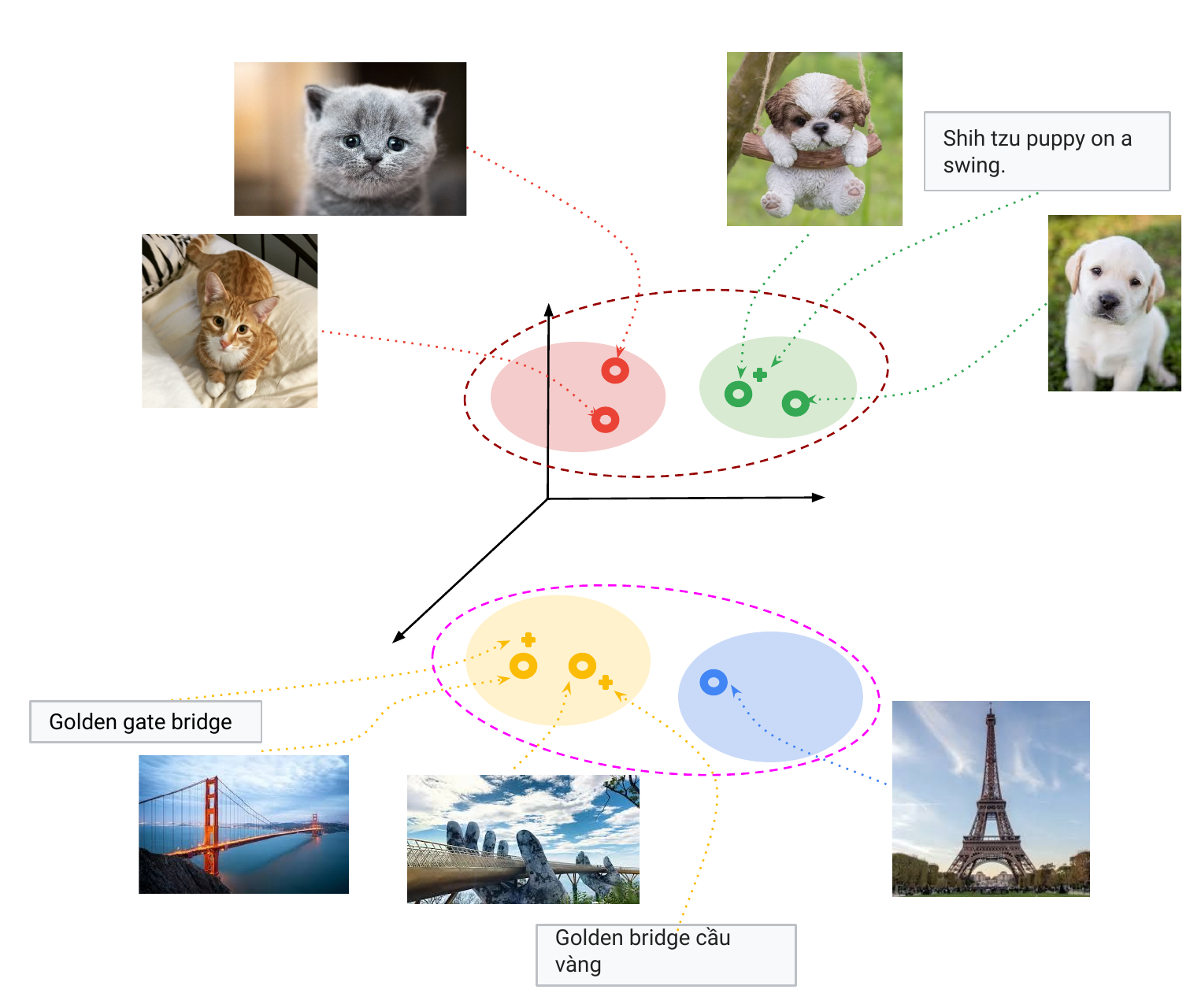}
\caption{An example universal embedding space. Cat, dog, bridge and tower classes are represented with red, green, yellow and blue colors respectively. Circles represent image embeddings and plus represent text embeddings.}
\label{fig:universal_embedding_space}
\end{figure}

For example, consider the embedding space corresponding to four classes: \textit{cat, dog, bridge, tower} in Figure~\ref{fig:universal_embedding_space}. In this space, the embeddings corresponding to an instance (e.g., image and text corresponding to \textit{Golden Gate bridge}) lie closest to each other. After that, the embeddings corresponding to a semantic class are closer to each other resulting in four clusters. Moreover,  the embeddings corresponding to \textit{cat} and \textit{dog} classes are closer to each other as they are semantically similar. Similarly, the embeddings corresponding to \textit{bridge} and \textit{tower} are closer to each other. This hierarchical semantic universal embedding space would be very useful in cross modal retrieval. When the \textit{dog} class is queried, \textit{dog} results would be retrieved first followed by \textit{cat} results.

The domain of visual semantic embeddings addresses this issue by incorporating semantic knowledge to image embeddings. Majority of these methods create an embedding space based on some external knowledge such as WordNet hierarchy or pre-trained word embeddings and projects image embeddings onto this fixed embedding space~\cite{frome2013devise, barz2019hierarchy}. Constraining the image embeddings in this manner might limit the performance of these embeddings. Moreover, there is no existing work on incorporating this semantic information to modalities beyond image to the best of our knowledge. Although these methods may be extensible to other modalities, the issue of media gap still needs to be addressed. 

We propose a novel method, HUSE, that addresses the above issues to learn Hierarchical Universal Semantic Embeddings. Unlike the previous ranking loss based methods that learn universal embeddings, HUSE projects images and text into a shared latent space by using a shared classification layer for image and text modalities. 
%
Unlike the visual semantic embedding methods that projects image embeddings onto a fixed semantic embedding space, HUSE allows the shared latent space to be completely independent of the semantic embedding space, while still having the similar semantic structure as that of semantic embedding space. HUSE incorporates semantic information by making the distance between any two universal embeddings to be similar to that of the distance between their class label embeddings in the semantic embedding space. 
%
%
HUSE achieves state-of-the-art retrieval, hierarchical precision, and classification results on UPMC Food-101 dataset.

In summary, the main contributions of the paper are:
\begin{enumerate}
\item A novel architecture that uses shared classification layer to learn universal embeddings.
\item Incorporation of semantic information into a universal embedding space.
\item A novel semantic embedding method that doesn't involve projection onto semantic embedding space.
\item State-of-the-art retrieval, hierarchical precision, and classification results on UPMC Food-101 dataset.
\end{enumerate}

\section{Related Work}
\label{sec:related_work}
This section briefly reviews the literature corresponding to universal embeddings, multimodal classification and semantic embeddings as our work is related to these areas.

\subsection{Universal Embeddings} The previous works on universal embeddings map images and text into a shared embedding space by maximizing correlation between related instances in the embedding space~\cite{andrew2013deep, gong2014multi, rasiwasia2010new, yan2015deep}. More recent works use ranking loss for universal embeddings~\cite{ye2018advise, wang2016learning, kiros2014unifying, faghri2017vse++, frome2013devise, carvalho2018cross}. Additional auxiliary tasks such as categorization~\cite{huang2018learning, rasiwasia2010new, salvador2017learning}, adversarial losses~\cite{gu2018look, wang2017adversarial, sarafianos2019adversarial} and multi-head self-attention ~\cite{song2019polysemous} are also used to align these embeddings better. 

Unlike these works, we learn universal embeddings using a shared hidden layer and classification objective. The image and text features extracted from pretrained networks are passed through their respective towers and the resulting embedding are L2 normalized. These normalized embeddings are classified by a shared hidden layer, which helps in mapping images and text into a shared embedding space. 

\subsection{Multimodal Classification} Previous works on multimodal classification can be primarly divided into early fusion, intermediate fusion and late fusion methods. Early fusion methods fuse the features from multiple modalities before passing into deep neural network~\cite{williams2018dnn, nawaz2018learning}. Intermediate fusion methods fuse (concatenation or cross modal attention) the features from the intermediate layers of deep networks processing different modalities and the fused features are classified~\cite{kiela2018efficient, arevalo2017gated}. Late fusion methods classify different modalities by different deep networks and the classification scores are fused by simple methods such as weighted average or by training another network on top of the classification scores ~\cite{wang2015recipe, narayana2018gesture, narayana2018interacting, narayana2018improving, narayana2019analyzing, narayana2019continuous}.

We employ the late fusion approach for multi-modal classification where the classification scores from image and text modalities are fused. However, unlike other late fusion approaches that have a separate classification layer for each modality, we use a shared hidden layer for all modalities. In addition to doing multimodal classification, this shared hidden layer helps in building a universal embedding space by clustering the embeddings corresponding to a class together.

\subsection{Semantic Embedding}
In computer vision literature, many prior works incorporated external knowledge to improve classification and image retrieval. As ImageNet classes are derived from WordNet, WordNet ontology is exploited for classification~\cite{chang2015large, verma2012learning, yan2015hd, zhao2011large} and image retrieval~\cite{deng2011hierarchical, li2017learning} tasks.
Frome~\etal incorporated prior world-knowledge by mapping pre-trained image embeddings onto word embeddings of class labels learned from text corpora~\cite{frome2013devise}. Barz and Denzler learned hierachy-based image embeddings by mapping images onto class embeddings whose pair-wise dot products correspond to a measure of semantic similarity between classes~\cite{barz2019hierarchy}. They computed class embeddings by a deterministic algorithm based on prior world-knowledge encoded in a hierarchy of classes such as WordNet. Juan~\etal learned image semantic embeddings based on the co-click information of the images~\cite{Juan:2019}.

In contrast to the above feature level approaches that maps image embeddings to word embeddings, our method learns a universal embedding space that is similar to the class level semantic embedding space in terms of distance. This gives the additional flexibility of having different dimensions compared to class level embedding space. 
\section{Method}
\label{sec:method}

\subsection{Problem Formulation}
\label{sec:problem}
Given a labeled set $\mathscr{D}$ that contains image-text-label triples $(p, q, y)$, where label $y$ is the class label corresponding to the image $p$ and text $q$, the objective is to learn a universal embedding space that achieves semantic understanding. Let $x^y$ correspond to the modality agnostic representation that represents either image $p$ or text $q$ with class label $y$.

Let $\phi_I(\cdot)$ represent a function that projects an image to a dense vector representing an embedding in universal embedding space and $\phi_T(\cdot)$ represent a function that projects a text to 
the universal embedding space. Let $\phi(\cdot)$ correspond to a modality agnostic projection function that projects $x$ to universal embedding space. $\phi(\cdot)$ corresponds to $\phi_I(\cdot)$ if $x$ is image and $\phi_T(\cdot)$ if $x$ is text. Let $d(\cdot)$ be the distance metric that measures the distance between two embeddings. Unless otherwise mentioned, we will use cosine distance in the rest of the paper.


The universal embedding space that achieves semantic understanding should have the following properties:
\begin{enumerate}
    
    \item \textbf{Class level similarity:} The distance between any two embeddings corresponding to the same class should be on average less than the distance between two embeddings corresponding to different classes. So,
    \begin{equation}
    d_{avg}(\phi(x_i^a), \phi(x_j^a)) < d_{avg}(\phi(x_m^a), \phi(x_n^b)),
    \end{equation}
    
    where $d_{avg}$ indicates average distance, computed over different (suitable) choices of point pairs.
    
    \item \textbf{Semantic similarity:} The embeddings corresponding to two different but semantically similar classes should be on average closer to each other than the embeddings corresponding to two semantically different classes. So,
    \begin{equation}
    d_{avg}(\phi(x_i^a), \phi(x_j^b)) < d_{avg}(\phi(x_m^a), \phi(x_n^c)),
    \end{equation}
    if $a$ and $b$ are semantically more similar classes than $a$ and $c$. 
    
    \item \textbf{Cross modal gap:} Cross modal gap makes the universal embedding learning challenging as the embeddings from different modalities have different distributions~\cite{peng2017overview}. Ideally, the distance between paired image and text should be close to zero as they correspond to the same instance. So, we would like to have
    \begin{equation}
    d(\phi_I(p_n),\phi_T(q_n)) \approx 0,
    \end{equation}
    for different points $n$ in the data.
\end{enumerate}

\subsection{Network Structure}
\label{sec:network}
\begin{figure}[ht]
 \includegraphics[width=\linewidth]{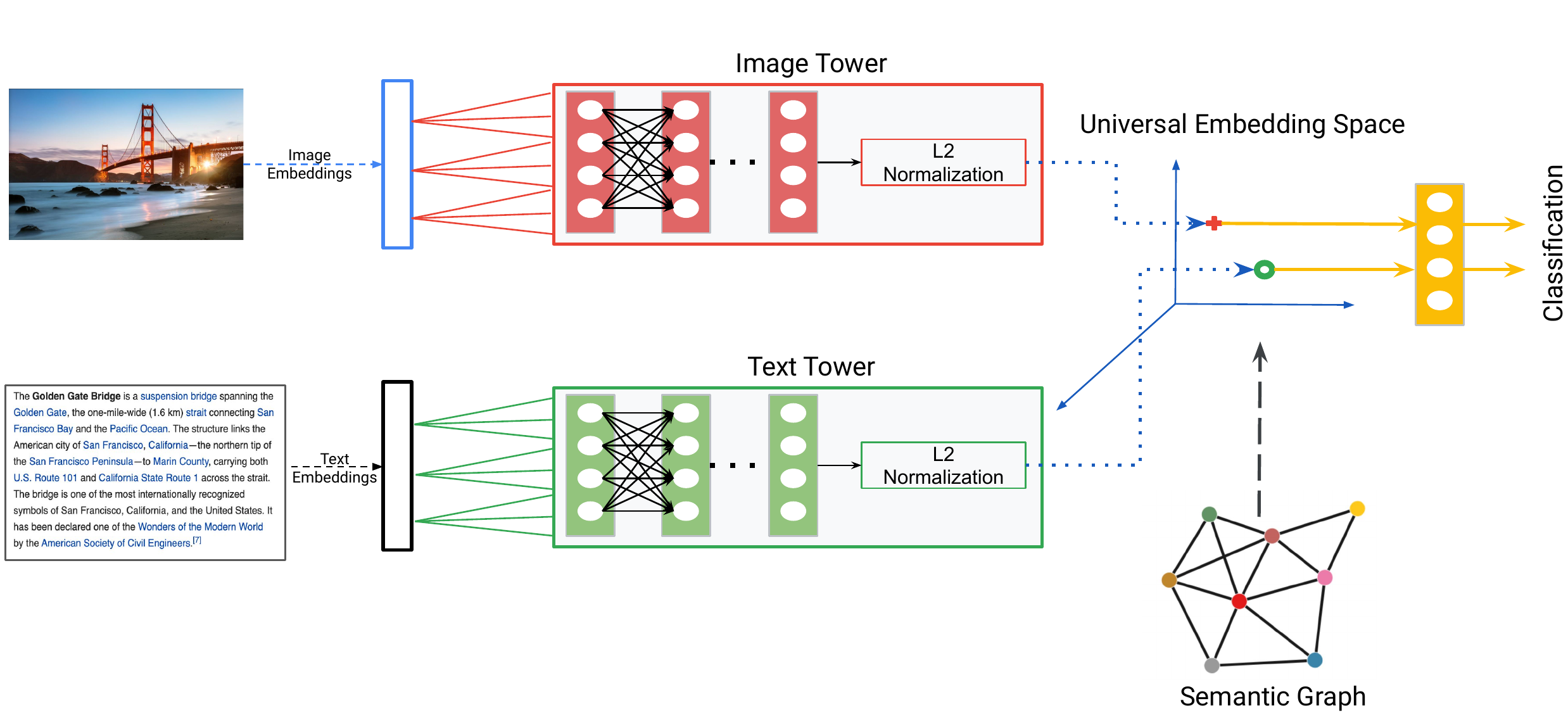}
\caption{Network architecture of HUSE. Image embeddings and text embeddings 
are passed through their respective towers to get universal embeddings. These embeddings are passed through a shared hidden layer. The semantic graph is used to regularize the universal embedding space.}
\label{fig:architecture}
\end{figure}

HUSE consists of an image tower that returns universal embeddings corresponding to an image. The image tower consists of multiple fully connected layers with RELU non-linearity and dropout. The last fully connected layer has $D$ hidden units and the embedding from the last layer is normalized to restrict the universal embedding space to a $D$-dimensional unit sphere. To make the training process simple, we extract image embeddings using pretrained backbone image network which are passed through an image tower to project them into a universal embedding space as shown in Figure~\ref{fig:architecture}. More formally, the backbone image network followed by the image tower corresponds to the image projection function $\phi_I(\cdot) \in \mathbb{R}^{D}$ that projects an image to $D$-dimensional universal embedding space. 

Similarly, HUSE model has a text tower consisting of multiple fully connected layers with RELU non-linearity and dropout. Similar to the image tower, the last fully connected layer has $D$ hidden units followed by $L_2$ normalization. Similar to images, we extract text embeddings using pretrained backbone text network and pass them through this text tower. More formally, the backbone text network followed by text tower corresponds to an text projection function $\phi_T(\cdot) \in \mathbb{R}^{D}$ that projects text to $D$-dimensional universal embedding space. 

The text tower and image tower can have different number of hidden layers and different hidden layer sizes. The only restriction is that the last layer of both towers should have $D$ hidden units followed by $L_2$ normalization.

The resulting embeddings from both image and text towers are passed through a shared fully connected layer of $K$ hidden units. This layer is used to classify the embeddings into $K$ classes.

\subsection{Semantic Graph}
\label{sec:semantic_graph}
We use the the text embeddings of class names extracted from a pretrained text model to construct a semantic graph. The distance between text embeddings corresponding to two words provide an effective method for measuring the semantic similarity of the corresponding words~\cite{pennington2014glove, cer2018universal}. As the text embedding space is semantically meaningful, previous works tried to project image embeddings into word embedding space~\cite{frome2013devise, li2017learning}.

Each class name is treated as a ``vertex" of the semantic graph and two class name's are connected by an ``edge". The cosine distance between two class name embeddings is treated as ``edge weight" so that two semantically similar classes have a lower edge weight compared to two semantically different classes. This semantic graph is used to regularize the universal embedding space.

More formally, we define the semantic graph as $G = (V,E)$, where $V = \{v_1, v_2, ..., v_K\}$ represents the set of $K$ classes and $E$ represent the edges between any two classes. Let $\psi(\cdot)$ represent the function that extracts embeddings of a class name. The adjacency matrix $A = \{A_{ij}\}_{i,j=1}^{K}$ of graph $G$ contains non-negative weights associated with each edge, such that:
\begin{equation}
A_{ij} = d(\psi(v_i), \psi(v_j)), 
\end{equation}
where $d$ is cosine distance as specified in Section~\ref{sec:problem}.

\subsection{Learning Algorithms}
To learn HUSE, the network architecture presented in Section~\ref{sec:network} is trained using the following loss:

\begin{equation}
\mathcal{L} = \alpha \mathcal{L}_{\mathrm{classification}} + \beta\mathcal{L}_{\mathrm{graph}} + \gamma \mathcal{L}_{\mathrm{gap}}
\end{equation} 

where $\mathcal{L}_{\mathrm{classification}},  \mathcal{L}_{\mathrm{graph}}, \mathcal{L}_{\mathrm{gap}}$ correspond to losses that will make the learned embedding space to have the three properties discussed in Section~\ref{sec:problem} and $\alpha, \beta \text{ and } \gamma$ correspond to weights to control the influence of these three losses. These three losses are discussed in more detail below.

\subsubsection{Class Level Similarity}

HUSE passes the embeddings from image tower and text tower through a shared fully connected layer and the model is trained using softmax cross entropy loss. This discriminative learning objective can cluster the embeddings corresponding to a class together and is easy to train than the usual metric learning objectives. Moreover, the network can also simultaneously do multimodal classification.

More formally, we denote the common fully connected layer as $\chi(\cdot)$ that calculates unnormalized log probabilities ($z$) of each class. For a training example $x_i$, these unnormalized log probabilites are calculated as: 

\begin{equation}
z_i = \chi(\phi(x_i)) = W^T \phi(x_i)+b 
\end{equation}


From the logits, the probability of each label $k \in \{1, . . . , K\}$ for a training example $x_i$ is calculated as 
\begin{equation}
p(k|x_i) = \frac{exp(z_i^k)}{\sum_{i=k}^{K}exp(z_i^k)},
\end{equation}

where $z_i^k$ is the unnormalized log probability for the $k$-th class for point $x_i$. Let $y_i$ correspond to the ground truth label of an example $x_i$. The softmax cross entropy loss is calculated as 

\begin{equation}
\mathcal{L}_{\mathrm{classification}} = -\frac{1}{N}\sum_{i=1}^{N} \sum_{k=1}^K I[y_i=k] log(p(k|x_i)),
\end{equation}
where $I$ is the indicator function.



\subsubsection{Semantic Similarity}
\label{sec:sem}

To make the learned universal embedding space semantically meaningful, where embeddings corresponding to two semantically similar classes are closer than the embeddings corresponding to two semantically different classes, we regularize the embedding space using the semantic graph discussed in Section~\ref{sec:semantic_graph}. This semantic graph regularization enforces additional constraint that the distance between any two embeddings is equal to the edge weight of their corresponding classes in semantic graph. As the semantically similar classes have smaller edge weight compared to semantically different classes, the regularization forces semantically similar classes to be closer to each other.

We use the following graph regularization loss to learn semantic similarity.

\begin{equation}
\label{equan:graph_loss}
\mathcal{L}_{\mathrm{graph}} = \frac{1}{N^2}\sum_{m=1}^{N}\sum_{n=1}^{N}(d(\phi(x_m^i), \phi(x_n^j))-A_{ij})^2
\end{equation}

However, it is hard to satisfy the constraint that all pairs of embeddings must adhere to the graph $G$. So, we relax this constraint by adding a margin so that the regularization is enforced on semantic classes which are closer than the margin and make other embedding pairs at least as large as the margin. 

\begin{equation}
    \sigma^{ij}_{mn}= 
\begin{cases}
    1 & \text{if } A_{ij}<\zeta \text{ and 
    } d(\phi(x_m^i), \phi(x_n^j))<\zeta\\
    0              & \text{otherwise}
\end{cases}
\end{equation}

After relaxing the constraint, the resulting loss is 

\begin{equation}
\label{equan:loss}
\mathcal{L}_{\mathrm{graph}} = \frac{1}{N^2}\sum_{m=1}^{N}\sum_{n=1}^{N}\sigma^{ij}_{mn}(d(\phi(x_m^i), \phi(x_n^j))-A_{ij})^2
\end{equation}

\subsubsection{Cross Modal Gap}
To reduce the cross modal gap, the distance between image and text  embeddings corresponding to the same instance should be minimized. The following loss function is used to achieve instance level similarity.

\begin{equation}
\mathcal{L}_{\mathrm{gap}} = \frac{1}{N}\sum_{n=1}^{N}d(\phi_I(p_n), \phi_T(q_n))
\end{equation}

\section{Experiments}
\label{sec: results}

We test our methodology on the UPMC Food-101 dataset. As HUSE incorporates semantic information derived from class labels when learning universal embeddings, we need a large multimodal classification dataset. UPMC Food-101 dataset~\cite{wang2015recipe} is a very large multimodal classification dataset containing around 101 food categories. The dataset has about 100,000 textual recipes and their associated images. Apple pie, baby back ribs, strawberry shortcake, tuna tartare are some examples of food categories in this dataset.

\subsection{Implementation Details}

This section discusses the visual and textual features used by HUSE, followed by the methodology used to construct semantic graph from class labels. Finally, we discuss the network parameters and training process.

\subsubsection{Feature Representation}
\begin{itemize}
\item \textbf{Visual:} We extract pretrained Graph-Regularized Image Semantic Embeddings (Graph-RISE) of size 64 from individual images~\cite{Juan:2019}. Graph-RISE embeddings are trained on 260 million images to discriminate 40 million ultra-fine-grained semantic labels using a large-scale neural graph learning framework. 



\item \textbf{Textual:} We use BERT embeddings\footnote{\href{https://github.com/google-research/bert}{https://github.com/google-research/bert}}~\cite{devlin2018bert} to obtain a representation of the text. Similar to Devlin~\etal~\cite{devlin2018bert}, we concatenate the embeddings from the last four layers for each token and then average all token embeddings for a document. In addition to the BERT encoding, we also calculate TF-IDF for the text features to provide a notion of importance for each word in relation to the whole corpus by following the pre-processing steps similar to~\cite{wang2015recipe}. 

As the number of tokens in each instance of UPMC Food-101 are far greater than the maximum tokens (512) that pre-trained BERT support, we extract 512 salient tokens from the text. To extract the salient tokens for a single example, we consider every sentence in the text as a separate document to build a corpus. We then use this corpus as the input to a TF-IDF model~\cite{rehurek_lrec} and the most important 512 tokens are extracted.

\end{itemize}

\subsubsection{Semantic Graph:} We construct a semantic graph based on the embeddings extracted from the class names as discussed in Section~\ref{sec:sem}. As class names often contain more than a single word (e.g., apple pie), we use Universal Sentence Encoder~\cite{cer2018universal} that provide sentence level embedding to extract embeddings of class names. To build the semantic graph, each class name is treated as a vertex and the cosine distance between universal sentence encoder embeddings of two class names is treated as edge weight. This semantic graph is used in the graph loss stated by Equation~\ref{equan:graph_loss}.

\subsubsection{Training Process}
The image tower consists of 5 hidden layers of 512 hidden units each and text tower consists of 2 hidden layers of 512 hidden units each. A dropout of 0.15 is used between all hidden layers of both towers. The network is trained using the RMSProp optimizer with a learning rate of 1.6192e-05 and momentum set to 0.9 with random batches of 1024 for 250,000 steps. These hyperparameters are chosen to maximize the image and text classification accuracies on the validation set of UPMC Food-101 dataset.

\subsection{Baselines}
\label{sec:baseline}
To test the effectiveness of HUSE, we compare it to the following state-of-the-art methods for cross-modal retrieval and visual semantic embeddings. As visual semantic methods are modeled only for image embeddings, we extend them to have textual embeddings as well. The input feature representation, number of hidden layers, hidden layer size and other training parameters of baselines are set similar to the HUSE model.
\begin{enumerate}

\item \textbf{Triplet:} This baseline uses the triplet loss with semi hard online learning to decrease the distance between embeddings corresponding to similar classes, while increasing the distance between embeddings corresponding to different classes~\cite{schroff2015facenet}.

\item \textbf{CME:} This method learns Cross-Modal Embeddings by maximizing the cosine similarity between positive image-text pairs, and minimizing it between all non-matching image-text pairs. CME uses additional classification loss for semantic regularization~\cite{salvador2017learning}.

\item \textbf{AdaMine:} AdaMine uses a double-triplet scheme to align instance-level and semantic-level embeddings~\cite{carvalho2018cross}.

\item \textbf{DeViSE*\footnote{* indicates that the original model is extended to support text embeddings}:} DeViSE maps pre-trained image embeddings onto word embeddings of class labels learned from text corpora~\cite{frome2013devise} using skip-gram language model. We extend DeViSE to support text by mapping pre-trained text embeddings onto word embeddings of class labels. To have a fair comparison, we use the class label embeddings used to construct the semantic graph instead of learning word embeddings of class labels using skip-gram language model as in the original paper.

\item \textbf{HIE*:} Hierarchy-based Image Embeddings maps images to class label embeddings and uses an additional classification loss~\cite{barz2019hierarchy}. We extend this model to support text embeddings by mapping text to class centroids and using an additional classification loss on text. Although the original paper deterministically calculates the class embeddings based on the hierarchy, we use the class label embeddings used to construct the semantic graph for fair comparison.

\item \textbf{HUSE-P:} This baseline has the similar architecture as that of HUSE and is meant to disentangle the architectural choices of HUSE from the graph loss. HUSE-P maps universal embedding space to class label embedding space by projecting image and text embeddings to class label embeddings similar to DeViSE* and HIE*.
HUSE-P uses the following projection loss instead of graph loss in equation~\ref{equan:loss}.
\begin{equation}
\mathcal{L}_{\mathrm{proj}} = \frac{1}{N}\sum_{n=1}^{N}d(\phi_I(p_n^i), \psi(v_i))+d(\phi_T(q_n^i), \psi(v_i))
\end{equation}
\end{enumerate}

We also compare the classification accuracy of HUSE to the previous published results by Wang~\etal~\cite{wang2015recipe} and Kiela~\etal~\cite{kiela2018efficient} on UPMC Food-101 dataset. As CME, HIE* and HUSE-P baselines discussed above also has classification layer, we also compare their classification accuracies to HUSE. In addition, we also use separate classification models on image and text modalities of UPMC Food-101 as another baseline for classification task.

\subsection{Retrieval Task}
The universal embedding space learned by HUSE can be used for both in-modal and cross-modal retrieval. This section quantifies the performance of HUSE on two in-modal retrieval tasks (Image2Image and Text2Text) and two cross-modal retrieval tasks (Image2Text and Text2Image) by comparing it to the baselines discussed in Section~\ref{sec:baseline}. Given a query (image or text), the retrieval task is to retrieve the corresponding image or text from the same class. For evaluation, we consider each item in the test set as a query (for instance, an image), and we rank the other candidates according to the cosine distance between the query embedding and the candidate embeddings. The results are evaluated using the recall percentage at top K (R@K), over all queries in the test set. The R@K corresponds to the percentage of queries for which the matching item is ranked among the top K closest results. 

\begin{table*}[]
\begin{tabular}{@{}ccccccccccccccccc@{}}
\toprule
        &  & \multicolumn{3}{c}{Image2Image} &  & \multicolumn{3}{c}{Image2Text} &  & \multicolumn{3}{c}{Text2Image} &  & \multicolumn{3}{c}{Text2Text} \\ \cmidrule{3-5} \cmidrule{7-9} \cmidrule{11-13} \cmidrule{15-17}
        
        &  & R@1        & R@5       & R@10      &  & R@1       & R@5       & R@10      &  & R@1       & R@5       & R@10      &  & R@1       & R@5       & R@10     \\ \midrule
Triplet &  & 0.469      & 0.685     & 0.746     &  & 0.282     & 0.541     & 0.632     &  & 0.171     & 0.410     & 0.539     &  & 0.484     & 0.705     & 0.783    \\
CME & &0.660&0.790&0.821& &0.430&0.552&0.592& &0.090&0.313&0.458& &0.802&0.880&0.901\\
AdaMine& &0.191&0.387&0.499& &0.042&0.160&0.266& &0.038&0.134&0.221& &0.215&0.350&0.447\\ \cdashlinelr{1-17}
DeViSE*&&0.656&0.793&0.830& &0.537&0.698&0.748& &0.220&0.438&0.539& &0.543&0.710&0.775\\
HIE*& &0.649&0.792&0.832& &0.590&0.712&0.751& &0.430&0.721&0.800& &0.786&0.870&0.897\\
HUSE-P& &0.668&0.794&0.828& &0.328&0.579&0.666& &0.689&0.826&0.853& &0.830&0.894&0.913\\
HUSE    &  & \textbf{0.685}      & \textbf{0.806}     & \textbf{0.835}     &  & \textbf{0.702}     & \textbf{0.763}     & \textbf{0.778}     &  & \textbf{0.803}     & \textbf{0.889}     & \textbf{0.903}     &  & \textbf{0.875}     & \textbf{0.917}     & \textbf{0.927}    \\
 \bottomrule
\end{tabular}
\caption{Retrieval performance of various methods on four retrieval tasks of UPMC Food-101 dataset. The entries above the dashed line correspond to the methods without semantic information and the entries below it correspond to methods that incorporate semantic information. The best value per column is set in bold font.}
\label{tab:r@k}
\end{table*}

Table~\ref{tab:r@k} shows the R@1, R@5, R@10 results on 4 retrieval tasks. The results show that HUSE is outperforming all the baselines across all retrieval tasks in all measures on UPMC Food-101 dataset. The more evident gains on cross-modal retrieval tasks compared to in-modal retrieval show that HUSE is able to address media gap more effectively than other methods. Moreover, the significant gains on R@1 metric compared to R@5, R@10 metrics show that HUSE is learning instance level semantics better than the other baselines. The first three entries in Table~\ref{tab:r@k} correspond to methods that don't include semantic information and the other entries correspond to semantic embedding methods. In general, the methods that uses the semantic information are performing better than the methods that are not using semantic information. This shows that the semantic information is important even when class level retrieval tasks are considered.

HUSE greatly outperforms all the baselines that doesn't use semantic information in all measures. The gains are even more significant on cross-modal retrieval tasks and this can be attributed to the shared classification layer and instance loss used by HUSE that forces the embeddings from both modalities to be more aligned. CME is the best performing model among the baselines that doesn't use semantic information and it's architecture is compartively more similar to HUSE as both models have a classification loss and instance loss. However, CME has separate classification layers for image and text and it's instance level loss minimizes the distance between embeddings corresponding to an instance while maximizing the distance between embeddings of different instances, even if they belong to the same class. This is different to the instance loss used by HUSE that only minimizes the distance between embeddings corresponding to an instance. AdaMine is the least performing baseline and it can be attributed to the double triplet loss the method is using. The instance level triplet loss pushes the embeddings corresponding to different instances away although they belong to same class, whereas the class level triplet loss tries to bring them closer. The polarity between these two loses makes AdaMine hard to optimize. 

The results also show that HUSE is significantly outperforming the baselines that use semantic information, although the performance difference is small compared to the methods that don't use semantic information. HUSE still outshines these methods on cross-modal retrieval tasks.  The semantic embedding baselines uses a fixed class label embedding space and learn mappings to project image and text to that space. HUSE, on the other hand, has the flexibility of learning an embedding space that is completely different from class label embedding space, but has the same semantic distances as the class label embedding space. This flexibility allows HUSE to learn better universal embedding space resulting in better retrieval performance.

\subsection{Semantic Quality}
As HUSE incorporates semantic information into the embedding space, this section evaluates the quality of the semantic embedding space learned by HUSE. To measure this, we employ a hierarchical precision@K (HP@K) metric similar to~\cite{frome2013devise} that measures the accuracy of model predictions with respect to a given semantic hierarchy. We create a taxonomy for the UPMC-Food 101 classes based on the WordNet ontology~\cite{fellbaum1998wordnet} and use it to calculate HP@K by generating a set of classes from the semantic hierarchy for each k, and computing the fraction of the model's k predictions that overlap with the class set. The HP@K values of all the examples in test set are averaged and are reported in Table~\ref{tab:hp@k}.
\begin{table*}[]
\begin{tabular}{@{}ccccccccccccccccc@{}}
\toprule
        &  & \multicolumn{3}{c}{Image to Image} &  & \multicolumn{3}{c}{Image to Text} &  & \multicolumn{3}{c}{Text to Image} &  & \multicolumn{3}{c}{Text to Text} \\ \cmidrule{3-5} \cmidrule{7-9} \cmidrule{11-13} \cmidrule{15-17}
        
        &  & H@2        & H@5       & H@10      &  & H@2       & H@5       & H@10      &  & H@2       & H@5       & H@10      &  & H@2       & H@5       & H@10     \\ \midrule
Triplet & &0.530&0.592&0.674& &0.383&0.493&0.604& &0.261&0.386&0.524& &0.520&0.581&0.669    \\
CME& &0.031&0.086&0.167& &0.033&0.077&0.162& &0.034&0.079&0.166& &0.031&0.085&0.167 \\
AdaMine&&0.164&0.252&0.376& &0.096&0.214&0.350& &0.084&0.179&0.309& &0.235&0.296&0.397\\  \cdashlinelr{1-17}
DeViSE*&&0.667&0.709&0.759& &0.683&0.742&0.799& &0.157&0.233&0.339& &0.350&0.412&0.503\\
HIE* &&0.682&0.718&0.768& &0.664&0.721&0.783& &0.522&0.607&0.698& &0.800&0.820&0.854\\
HUSE-P&& 0.689&0.722&0.768&&0.656&0.729&0.776&&0.573&0.632&0.763&&0.825&0.836&0.871\\
HUSE &&\textbf{0.705}&\textbf{0.740}&\textbf{0.787}& &\textbf{0.741}&\textbf{0.784}&\textbf{0.831}& &\textbf{0.824}&\textbf{0.848}&\textbf{0.878}& &\textbf{0.884}&\textbf{0.897}&\textbf{0.916 }      \\
 \bottomrule
\end{tabular}
\caption{Hierarchical Precision of various methods on four retrieval tasks of  UPMC Food-101. 
}
\label{tab:hp@k}
\end{table*}
HUSE outperforms all baselines by a significant margin on HP@K metric showing that HUSE is learning better semantic embedding space than other baselines. Even when the performance of the best method in each retrieval task is compared to HUSE, the performance improvements shows the superiority of HUSE in integrating semantic information into the embedding space. Moreover, these gains are even more prominent in cross-modal retrieval tasks (0.055 to 0.251) than in-modal retrieval taks (0.016 to 0.061). This performance improvements can be attributed to the fact that HUSE doesn't confine the universal embedding space to class label embedding space. Among three baselines that use the semantic information, HUSE-P is incorporating more semantic information than the other two methods as measured by HP@K. The only difference between HUSE and HUSE-P is that HUSE-P uses projection loss similar to DeViSE* and HIE*.
As all of these three baselines maps image and text embeddings to class label space, the architectural choice of using a shared classification layer and an instance loss are effectively reducing the media gap resulting in better performance.


\subsection{Classification Task}
As HUSE is trained with classification objective, it is natural to apply the model for classification task. For an image, text pair, HUSE returns separate classification scores for image and text. These softmax scores are fused together using simple weighted averaging. Table~\ref{tab:class} reports image, text and fusion classification accuracies of HUSE and other baselines along with the previous reported results on UPMC Food-101 dataset. 

HUSE achieved an accuracy of 92.3\% on UPMC Food-101 dataset, outperforming the previous state-of-the-art by 1.5\%. Moreover, the previous state-of-the-art model used complex gated attention method for fusing image and text channels~\cite{kiela2018efficient}. We, on the other hand, use a simple weighted averaging to fuse softmax scores from image and text channels. At the individual channel level, HUSE's image accuracy outperformed the previous best by a large margin (17.1\% gain) while performing slightly worse (0.7\% loss) on text channels.

\begin{table}
  \centering
  \begin{tabular}{@{}ccccccc@{}}
    \toprule
                                   && Image                   & & Text                                && Fusion               \\ \midrule
    Wang~\etal ~\cite{wang2015recipe}                           && 40.2                     && 82.0                                         && 85.1                 \\
    Kiela~\etal ~\cite{kiela2018efficient}                          && 56.7                     && \textbf{88.0}                                        && 90.8                 \\
    Separate Models                     && 72.4                     &&      87.2                                         &&        91.9              \\
    CME&& 72.4 && 78.8 && 88.1\\
    HIE*&&73.5 && 80.2&&88.5\\
    HUSE-P &&73.1 &&83.9 && 89.6 \\
    HUSE            && \textbf{73.8}                         &&      87.3                    &&    \textbf{92.3}                                       \\ \bottomrule
  \end{tabular}
  \caption{Classification accuracy of UPMC Food-101 dataset.}
  \label{tab:class}
\end{table}

However, the state-of-the-art classification results on this dataset can't be entirely attributed to HUSE. This is because we are using different image (Graph-RISE~\cite{Juan:2019}) and text (TFIDF+BERT~\cite{devlin2018bert}) embeddings compared to the previous state-of-the-art. To disentangle the contribution of embeddings and HUSE architecture on the classification accuracies, we trained ``separate" image classification and text classification models using the same hyperparameters as HUSE. These models are simple classification models without any semantic regularization and have  the hidden layers and hidden dimensions similar to image and text tower of HUSE. We see that these separate models achieve the fusion accuracy of 91.9\% outperforming the previous best by 1.1\%. The HUSE architecture further improves this score by another 0.4\%. The majority of the gains HUSE achieved on image channel can be attributed to the better image embeddings, yet HUSE improved this further by another 1.4\%. On the text channel, the performance of HUSE and the separate classification model are similar. Unlike separate classification models, HUSE imposes additional constraints on image and text channels by making them share a single classification layer and on the embedding space by including semantic information. However, these constraints didn't regress the classification accuracies of HUSE, but improved them.

Table~\ref{tab:class} also reports the classification accuracy of HUSE-P, where we replace the graph regularization loss with the projection loss to map the image and text embeddings onto the class label embeddings. We see that the performance of HUSE-P is inferior to HUSE. More interestingly, we see the performance regression on text and fusion accuracies compared to the baseline of separate models. These results show that mapping the universal embedding space to class embedding space degrades the resulting embeddings, whereas constraining the universal embedding space to have the same semantic distance as class embedding space improves the performance.


Based on these results, we can say that constraining the embedding space for cross-modal retrieval will decrease the classification performance compared to unconstrained classification models. Adding semantic information to the embedding space boosts the classification performance. Instead of mapping the universal embedding space to class label embedding space, allowing the universal embedding space to have the same semantic distance as the class label embedding space significantly improves the classification performance beyond the unconstrained classification models.

\section{Conclusion}

We proposed a novel architecture, HUSE, to learn a universal embedding space that incorporates semantic information. Unlike the previous methods that maps image and text embeddings to a constant class label embedding space, HUSE learns a new universal embedding space that still has the same semantic distance as the class label embedding space. These less constrained universal embeddings outperformed several other baselines on multiple retrieval tasks. Moreover, the embedding space learned by HUSE has more semantic information than the other baselines as measure by HP@K metric. A shared classification layer used by HUSE for both image and text embeddings and the instance loss reduced the media gap and resulted in superior cross-modal performance. Moreover, HUSE also achieved state-of-the-art classification accuracy of 92.3\% on UPMC Food-101 dataset outperforming the previous best by 1.5\%.



\bibliography{references}
\bibliographystyle{plain}

\end{document}